\documentclass[11pt]{article}

\usepackage[preprint]{acl}

\usepackage{times}
\usepackage{latexsym}
\usepackage{amsmath}
\usepackage{amsthm}
\usepackage{amssymb}
\usepackage{multirow}
\usepackage{booktabs}
\usepackage[table]{xcolor}

\usepackage[T1]{fontenc}

\usepackage[utf8]{inputenc}

\usepackage{microtype}

\usepackage{inconsolata}

\usepackage{graphicx}

\theoremstyle{plain}
\newtheorem{theorem}{Theorem}

\theoremstyle{definition}
\newtheorem{assumption}{Assumption}

\title{MIND: From Passive Mimicry to Active Reasoning through Capability-Aware Multi-Perspective CoT Distillation}

\author{
    Jin Cui\textsuperscript{$*1$}, 
    Jiaqi Guo\textsuperscript{$*2$}, 
    Jiepeng Zhou\textsuperscript{$*3$}, 
    Ruixuan Yang\textsuperscript{$1$}, \\
    \textbf{Jiayi Lu}\textsuperscript{$1$}, 
    \textbf{Jiajun Xu}\textsuperscript{$4$},
    \textbf{Jiangcheng Song}\textsuperscript{$1$},
    \textbf{Boran Zhao}\textsuperscript{$4$},
    \textbf{Pengju Ren}\textsuperscript{$\dagger1$} \\[1mm]
    \textsuperscript{$1$}State Key Laboratory of Human-Machine Hybrid Augmented Intelligence,\\
    National Engineering Research Center for Visual Information and Applications,\\
    and Institute of Artificial Intelligence and Robotics, Xi'an Jiaotong University\\
    \textsuperscript{$2$}Nankai University, 
    \textsuperscript{$3$}The Hong Kong University of Science and Technology(Guangzhou)\\
    \textsuperscript{$4$}School of Software Engineering, Xi'an Jiaotong University\\[-1mm]
    {\tt\small andycui@stu.xjtu.edu.cn, \{boranzhao, pengjuren\}@xjtu.edu.cn}
}

\begin{document}
\maketitle{
    \renewcommand{\thefootnote}{\fnsymbol{footnote}}
    \footnotetext[1]{Equal contribution.} \footnotetext[2]{Corresponding author.}   
}
\begin{abstract}
\vspace{-2mm}
While Large Language Models (LLMs) have emerged remarkable capabilities in complex tasks through Chain-of-Thought reasoning, practical resource constraints have sparked interest in transferring these abilities to smaller models. However, achieving both domain performance and cross-domain generalization remains challenging. Existing approaches typically restrict students to following a single golden rationale and treat different reasoning paths independently. Due to distinct inductive biases and intrinsic preferences, alongside the student's evolving capacity and reasoning preferences during training, a teacher's "optimal" rationale could act as out-of-distribution noise. This misalignment leads to a degeneration of the student's latent reasoning distribution, causing suboptimal performance.  To bridge this gap, we propose MIND, a capability-adaptive framework that transitions distillation from passive mimicry to active cognitive construction. We synthesize diverse teacher perspectives through a "Teaching Assistant" network. By employing a \textit{Feedback-Driven Inertia Calibration mechanism}, this network utilizes inertia-filtered training loss to align supervision with the student’s current adaptability, effectively enhancing performance while mitigating catastrophic forgetting. Extensive experiments demonstrate that MIND achieves state-of-the-art performance on both in-distribution and out-of-distribution benchmarks, and our sophisticated latent space analysis further confirms the mechanism of reasoning ability internalization.
\end{abstract}
\vspace{-5mm}
\section{Introduction}
 \vspace{-2mm}

Large Language Models (LLMs) have exhibited remarkable emergent capabilities in solving complex reasoning tasks \citep{wei2022emergent, bubeck2023sparks}. As an emergent ability, Chain-of-Thought (CoT) reasoning empowers models with exceptional performance by decomposing intricate problems into intermediate steps \citep{kojima2022large}. However, these capabilities typically mandate massive parameter scales \citep{wei2022chain}, rendering deployment in resource-constrained environments prohibitively expensive. To bridge this gap, CoT Distillation has emerged as a promising paradigm to transfer the reasoning prowess of LLMs into compact Student Models (SLMs) using teacher rationales as supervision \citep{Ho2022LargeLM, magister2023teaching, hsieh2023distilling}. Despite progress, we identify several limitations in current approaches that hinder the cultivation of robust reasoning:

\begin{figure}
    \centering
    \includegraphics[width=1\linewidth]{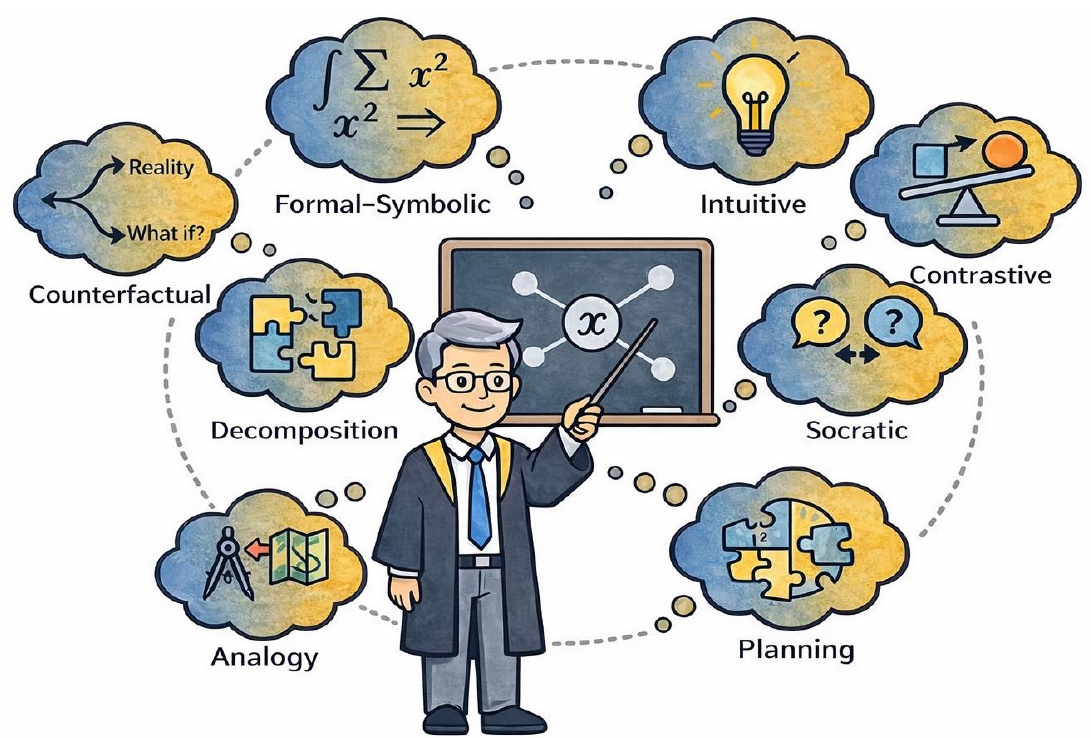}
    \caption{\textbf{Reasoning Perspectives inherent in LLMs.} We only demonstrate the distinctive ones used to clearly demonstrate our distillation method. Selectively using a subset for efficiency does not sacrifice performance.}
    \vspace{-7mm}
    \label{fig:perspectives}
\end{figure}
 
1) \textit{Distribution Collapse via Single-Path Rigidity.} 
Although teacher LLMs provide diverse reasoning trajectories (Figure \ref{fig:perspectives}), SLMs with limited capacity often fail to capture such complex multi-modal reasoning distributions. Rigid single-path supervision forces SLMs to average over diverse modes, causing the loss of strategic diversity to adaptively switch strategies for different questions \citep{Ho2022LargeLM, magister2023teaching, chen2023mcc} and brittle generalization.

2) \textit{Neglect of Structural Synergy among Reasoning Paths.} 
While recent works leverage multiple reasoning paths to improve reliability \citep{chen2023mcc, li2024mode}, they typically aggregate results through voting or ranking, implicitly assuming path independence, neglecting the structural synergy where strategies interact to resolve ambiguities and complement partial information \citep{ainsworth2006deft}. The absence of this mutual reinforcement leads to suboptimal supervision.


 
3) \textit{Misalignment from Static Supervision.}
Most critically, traditional "one-size-fits-all" supervision neglects the intrinsic teacher-student cognitive gap. Due to distinct inductive biases and preferences \citep{chen2025unveiling, jiang2025teach}, a teacher's "optimal" reasoning may act as out-of-distribution noise to the student that creates high variance gradients. Furthermore, we discovered that the student's learning capacity and reasoning preference evolve dynamically during training (e.g., preferring explicit step-by-step guidance over abstract leaps in early training steps) \citep{lin2025perspective}. This misalignment forces the student to learn patterns incompatible with their current capability, resulting in inefficient training and reasoning hallucinations.


To transition SLMs from passive mimicry to active reasoning, we propose Capability-Adaptive \textbf{M}ulti-Perspect\textbf{I}ve Chai\textbf{N}-of-Thought \textbf{D}istillation (\textbf{MIND}), a dynamic framework that harmonizes diverse reasoning patterns with student-centric adaptive supervision. 
Inspired by the distinct stylistic signatures of teacher LLMs, 
MIND constructs a \textit{multi-perspective corpus} to capture varied cognitive reasoning strategies. Through a \textit{dynamic fusion mechanism}, our framework enables SLMs to explicitly internalize these patterns, empowering them to flexibly synthesize appropriate strategies during inference, marking a significant leap from the monotonic reasoning of traditional methods.
To bridge the capability gap, we introduce Meta-Gating Network (MetaNet), a "Teaching Assistant" that facilitates cognitive alignment through a \textit{Feedback-Driven Inertia Calibration mechanism}. By leveraging inertia-filtered training loss as a proxy for student adaptability, MetaNet dynamically recalibrates fusion weights to direct supervision toward the most compatible paths. This capability-aligned process mitigates hallucinations and maximizes training efficiency even with minimal samples.

We pioneer framing distillation as a capability-aware cognitive construction process, departing from conventional monotonic supervision. Extensive evaluations on In-Distribution(ID) and Out-of-Distribution(OOD) benchmarks demonstrate that MIND achieves SOTA performance and mitigates catastrophic forgetting. These results suggest that our approach evolves SLMs from rote "task-takers" into genuine "thinkers" capable of universal reasoning.
Our main contributions are summarized as follows:
\vspace{-2mm}
\begin{enumerate}
    \item We propose MIND, a capability-aware multi-perspective framework that transitions distillation from mimicry to active cognitive construction, effectively enhancing performance.
    \vspace{-7mm}
    \item We introduce a "Teaching Assistant" to align teacher supervision with the student's evolving capability, effectively mitigating hallucination and catastrophic forgetting.
    \vspace{-2.5mm}
    \item Extensive experiments show MIND achieves SOTA performance across both ID (+3.27\%) and OOD (+7.53\%) benchmarks compared to previous SOTA. We also provide a rigorous latent space analysis to empirically verify that MIND empowers students to internalize topologically separable reasoning primitives rather than merely memorizing surface templates.
\end{enumerate}
\vspace{-4mm}

\begin{figure*}
    \centering
    \includegraphics[width=1\linewidth]{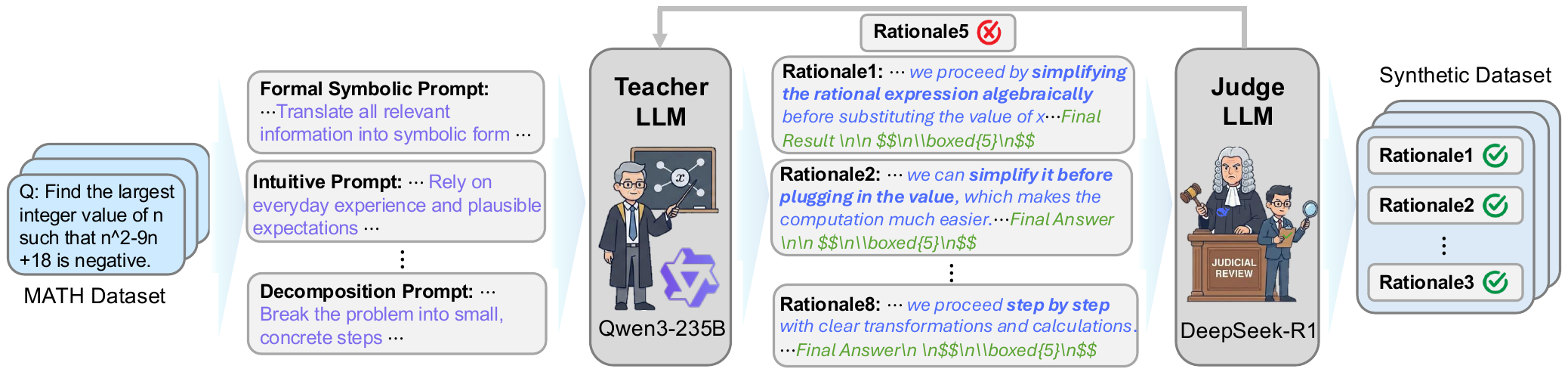}
    \caption{\textbf{Demonstration of the dataset construction pipeline.} We adopt a Judge LLM to identify reasoning with poor quality or leading to wrong answers, and maintain the consistency between rationales and the correct answer.}
    \vspace{-4mm}
    \label{fig:Dataset}
\end{figure*}


\section{Related Work}
\vspace{-1mm}
\subsection{Chain-of-Thought Capability in LLMs}
Large Language Models (LLMs) have demonstrated remarkable emergent capabilities \citep{wei2022emergent, brown2020language, nye2021show}., 2021], primarily realized by the Chain-of-Thought (CoT) reasoning paradigm \citep{wei2022chain, kojima2022large}. By decomposing complex problems into sequential intermediate steps, this strategy enables models to tackle intricate tasks that were previously intractable for standard answer-oriented inference \citep{wei2022emergent, chowdhery2023palm, imani2023math}. This approach has been shown to substantially enhance performance in both zero-shot and few-shot settings, eliciting deductive reasoning abilities that were previously latent \citep{kojima2022large, wang2023selfconsistency}. However, these emergent reasoning capabilities are strongly correlated with model scale; smaller language models (SLMs) often fail to spontaneously generate coherent reasoning chains or exhibit diminished performance compared to their larger counterparts \citep{magister2023teaching, fu2023specializing}, which evidences the importance of CoT distillation to explicitly endow SLMs with structured reasoning capabilities.
\subsection{Distill Reasoning Capabilities into SLMs}
While CoT Distillation effectively transfers reasoning capabilities from massive LLMs to compact Student Models (SLMs) \citep{Ho2022LargeLM, magister2023teaching, hsieh2023distilling}, recent studies emphasize that the diversity of reasoning paths and the granularity of rationales often impact distillation quality more than the teacher's raw accuracy. Meanwhile, some studies highlight significant limitations in this "imitation-only" approach as it often leads to spurious correlations between questions and answers, restricting generalization to out-of-distribution (OOD) tasks \citep{dai2024improve, feng2024teaching, wang2023scott, chen2023mcc}.

Crucially, existing methods struggle to harmonize diverse reasoning patterns with dynamic student adaptation. Multi-expert frameworks like \citep{li2024mode} rely on fixed, task-level weighting, which enforces a static reasoning mode and precludes the intra-step cognitive shifting required for complex problems. Similarly, recent adaptive approaches often rigidify the learning process: \citep{jiang2025teach} strictly segregates intuition from expression, while \citep{chenglin2024mixed} treats distinct strategies as competing options to be linearly weighted rather than complementary perspectives to be synthesized. Furthermore, these methods typically depend on external heuristics or predefined schedules, ignoring the student's internal cognitive state. Although contrastive sampling \citep{wang2025qcrd} attempts to mitigate this, it incurs prohibitive computational overhead. In contrast, our work is the first to explicitly synthesize a diverse reasoning repertoire synchronized with the student's real-time adaptability, achieving robust performance without auxiliary overhead.

\section{Method}

\begin{figure*}
    \centering
    \includegraphics[width=0.9\linewidth]{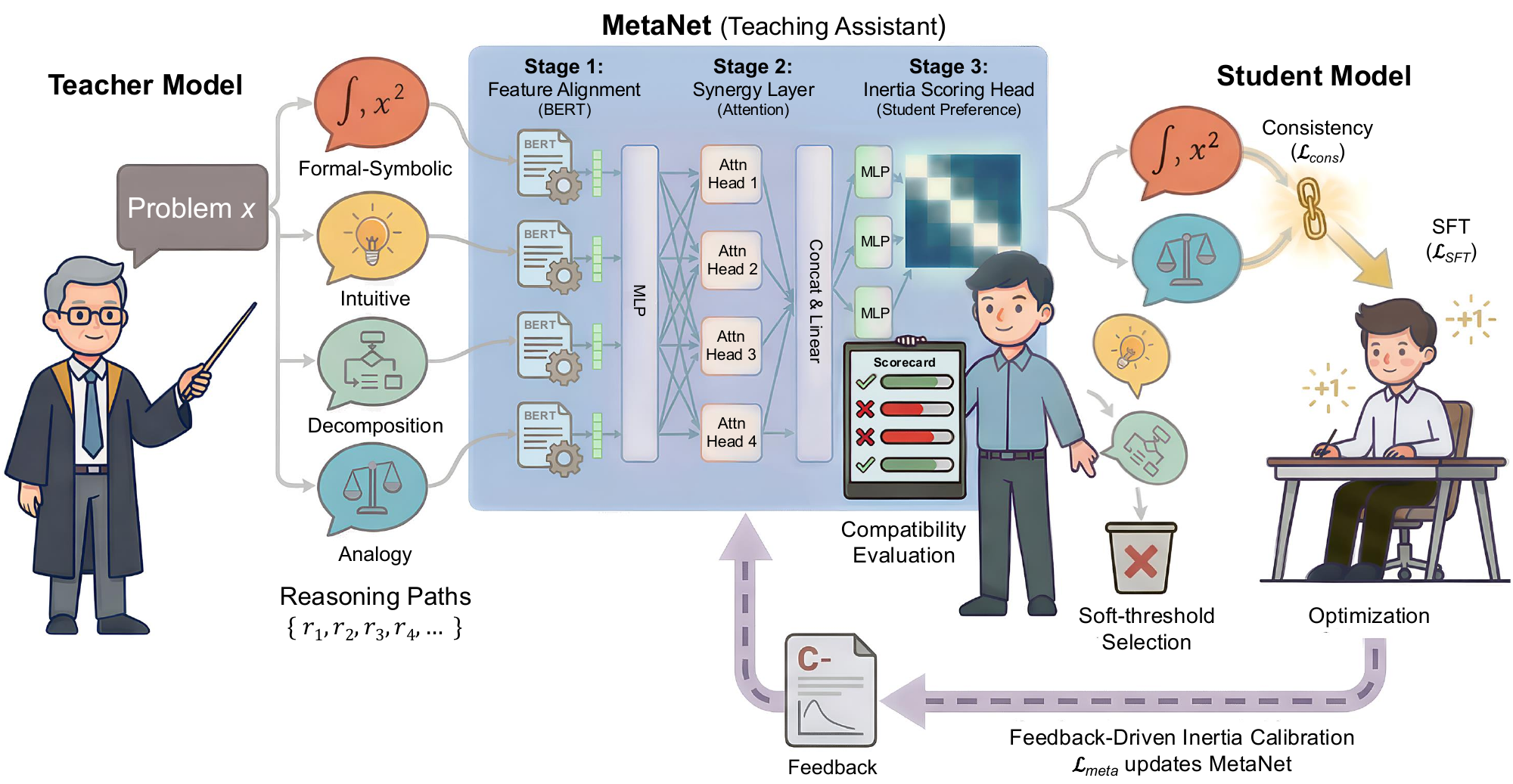}
    \vspace{-3mm}
    \caption{\textbf{Overview of our method MIND.} (1) We first prompt the teacher model to generate multi-perspective rationales. (2) Then, we warm up the MetaNet for a few steps on the acquired dataset and recalibrate it using the student's performance feedback. (3) We train the student model with SFT supervision and consistency regularization.}
    \vspace{-4mm}
    \label{fig:main-figure}
\end{figure*}

\subsection{Dataset Construction}
\label{sec:Dataset Construction}


\textit{Cognitive Perspectives and Prompting.} 
Empirically, we observe that teacher rationales exhibit clear semantic separability (Figure~\ref{fig:latent_state_visualization}(c)), reflecting distinct cognitive modes rather than a chaotic distribution. Leveraging this, we synthesize eight orthogonal Cognitive Perspectives (Figure \ref{fig:perspectives}) to maximize representational distinctiveness and coverage. To instantiate these abstract patterns, we utilize the MATH dataset \citep{hendrycks2021measuring} as our foundational corpus, chosen for its heterogeneous domains (e.g., algebra, geometry, number theory) that naturally necessitate distinct reasoning depths. Formally, for each sample $(x, y) \in \mathcal{D}_{\text{MATH}}$, we design perspective-specific prompts $\mathcal{P} = \{p_k\}_{k=1}^8$ to explicitly instruct the teacher (Qwen3-235B) to employ the $k$-th strategy as detailed in Figure~\ref{fig:Dataset}. This yields a multi-perspective dataset where each sample consists of candidate reasoning paths and predictions: $(x, \{r_k\}_{k=1}^8, \{\hat{y}_k\}_{k=1}^8)$.


\textit{Quality Filtering and Difficulty Stratification.}
To ensure corpus quality, we implement a rigorous post-processing pipeline. First, we discard traces where the prediction $\hat{y}_k$ deviates from the ground truth $y$, retaining only valid samples with correct results. Second, to avoid overfitting to trivial patterns and focus on complex reasoning, we downsample simple problems (MATH Levels 1-2) while preserving all challenging instances (Levels 3-5). The resulting dataset $\mathcal{D}_{MultiPers}$ features a balanced distribution across difficulties and perspectives.


\subsection{Latent Space Visualization}
\label{sec:Latent Space Analysis of Reasoning Perspectives}



\begin{figure}
    \centering
    \includegraphics[width=1\linewidth]{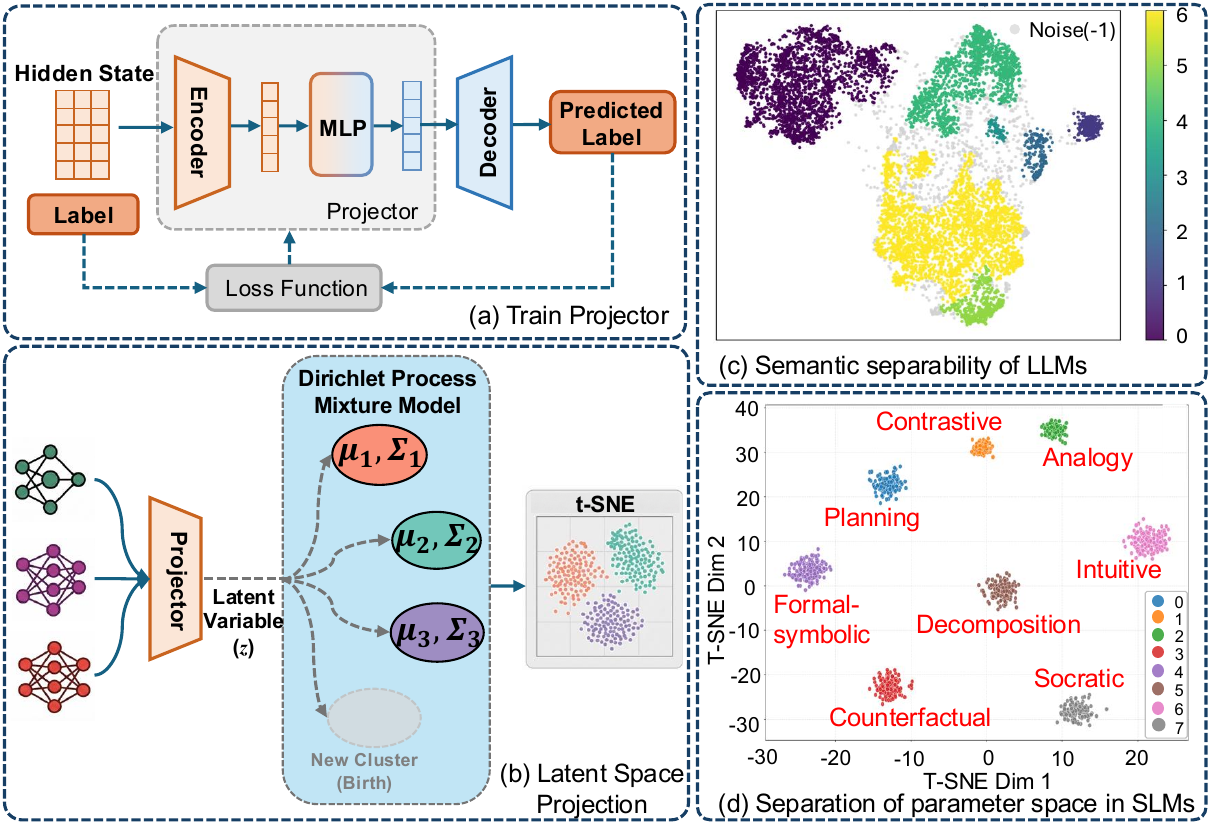}
    \caption{\textbf{Methodology and results of the Latent Space Visualization Analysis.} (a) and (b)demonstrate the mechanism, (c) illustrates the semantic level separation of LLM's reasoning paths, (d) provides a rigorous visualization of the differentiation of specifically trained students' latent space.}
    \vspace{-6mm}
    \label{fig:latent_state_visualization}
\end{figure}

To rigorously quantify whether the acquired reasoning patterns represent fundamentally different internal representations or merely superficial lexical variations, we visualize the student's latent reasoning manifold (Figure~\ref{fig:latent_state_visualization}(a)(b)).

We train an encoder to project the last-layer hidden states from eight "specialist" students (each distilled exclusively on a specific perspective) into a low-dimensional reasoning manifold $z$. To rigorously organize and interpret these representations, we incorporated a Dirichlet Process Mixture Model (DPMM) to govern the structure of the latent space. Unlike parametric models with fixed cluster constraints, the non-parametric DPMM spontaneously infers the underlying structure, allowing the active components to be determined solely by the observed data distribution. We then visualize the resulting topology of this latent space by t-SNE.

As shown in Figure~\ref{fig:latent_state_visualization}(d), the visualization reveals a highly structured manifold where the eight specialists form distinct, compact clusters with clear decision boundaries. This topological separation confirms that the distillation process has successfully imprinted stable, distinguishable cognitive signatures onto the student's parameter space. It demonstrates that distinct reasoning perspectives correspond to specific activation patterns, indicating the internalization of underlying mechanisms rather than surface-level template memorization.

\subsection{Capability-Aware Perspective Fusion} 
Having established that distinct reasoning perspectives form separable topological clusters within the student's latent space (Section \ref{sec:Latent Space Analysis of Reasoning Perspectives}), the subsequent challenge is to dynamically synthesize these isolated capabilities.

\subsubsection{Meta-Gating Network Architecture} 
We introduce the Meta-Gating Network (MetaNet) acting as a dynamic "Teaching Assistant" to evaluate the compatibility of each teacher-generated reasoning path with the student's current learning state. Formally, given an input question $x$ and a set of $K$ candidate reasoning paths $\mathcal{R} = \{r_k\}_{k=1}^K$, MetaNet predicts compatibility scores $\{s_k\}_{k=1}^K$ via three key components (Figure \ref{fig:main-figure}):

\noindent \textbf{1. Feature Alignment:} A frozen Sentence-BERT first encodes the question and reasoning paths into semantic embeddings $h_x$ and $h_{r_k}$. A learnable projection layer then fuses these features into a unified latent space: $e_k = \text{MLP}_{align}([h_x; h_{r_k}])$.

\noindent \textbf{2. Perspective Synergy:} To capture the structural dependencies among distinct strategies, we employ a Multi-Head Self-Attention layer. The layer aggregates information across perspectives, enhancing the representation of mutually reinforcing strategies: $Z = \text{SelfAttn}(E)$, where $E = [e_1, \dots, e_K]$.

\noindent \textbf{3. Adaptive Scoring:} We utilize $K$ independent MLP heads, parameterized by $\phi_k$, to function as a parametric "Inertia Matrix." By filtering short-term training variance, these heads effectively encode the student’s accumulating preferences for specific reasoning styles directly into their parameters. The output, $s_k = \text{MLP}_{score}^{(\phi_k)}(z_k)$, provides a stabilized estimation of the student's dynamic adaptability.

\subsubsection{Feedback-Driven Inertia Calibration} 

Ideally, supervision weights should prioritize patterns aligning with the student's evolving capability. A naive approach might directly utilize the student's real-time training loss to weight different paths. However, instantaneous loss is noisy and fails to reflect long-term cognitive evolution. To address this, we propose a \textit{Feedback-Driven Inertia Calibration Mechanism}.

\noindent \textbf{MetaNet Calibration.} Instead of constantly reacting to transient loss fluctuations, we explicitly align MetaNet's predictions with the student's evolving capacity using the student's real-time training loss, $\mathcal{L}_{real} = [-\log P(r_k|x; \theta)]_{k=1}^K$, as a ground-truth proxy for "learnability." After warming-up for a few steps, MetaNet is updated simultaneously with the student via a ListNet-based ranking loss to minimize the divergence between predicted scores $s$ and actual performance:

\vspace{-6mm}
\begin{equation}
\mathcal{L}_{\mathrm{meta}} = D_{\mathrm{KL}}\bigl( \pi_{\tau}(s) \,\Vert\, \pi_{\tau}(-\mathcal{L}_{\mathrm{real}}) \bigr)
\end{equation}

where $\pi_{\tau}=Softmax(\frac{s}{\tau})$ denotes the tempera-ture-scaled score. To ensure stability, we implement a differential update schedule where MetaNet employs a lower learning rate and higher gradient accumulation steps than the student. This induces a necessary optimization hysteresis, allowing the MetaNet to function as a stable anchor that filters stochastic noise while preserving the student’s parameter plasticity during early exploration and preventing coupled oscillation. 

\noindent \textbf{Consistency-Regularized Supervision.} Based on the compatibility scores $s$ predicted by MetaNet for all $K$ perspectives. To focus the student's limited capacity on high-value paths and filter outlier noise, we adaptively select a subset of high-confidence perspectives $\mathcal{I}_{dyn}$ that fall within a tolerance parameter $\beta$ of the optimal score:$$\mathcal{I}_{dyn} = \{k \mid s_k \geq \max(\mathbf{s}) * \beta\}$$ Fusion weights are then normalized exclusively within this valid subset:

\vspace{-6mm}
$$\alpha_k = \frac{\exp(s_k/\tau_{student})}{\sum_{j \in \mathcal{I}_{top}} \exp(s_j/\tau_{student})}, \quad \forall k \in \mathcal{I}_{top}$$

The student optimization objective $\mathcal{L}_{student}$ comprises two terms: a preference-weighted SFT loss and a pairwise consistency regularization loss.

\noindent \textbf{(1) Preference-Weighted SFT ($\mathcal{L}_{SFT}$):} We maximize the likelihood of the selected reasoning paths within $\mathcal{I}_{dyn}$, weighted by their compatibility $\alpha_k$. This ensures the student primarily learns from the perspectives that align best with its current cognitive state:

\vspace{-4mm}
\begin{equation}
\mathcal{L}_{SFT} = \sum_{k \in \mathcal{I}_{dyn}} \alpha_k \cdot \mathcal{L}_{CE}(r_k|x; \theta) 
\end{equation}

\noindent \textbf{(2) Pairwise Consistency Regularization ($\mathcal{L}_{cons}$):}  To prevent reasoning fragmentation, we enforce logical consistency among the selected perspectives with a consensus that valid reasoning paths, despite their diverse trajectories, should converge to consistent answer distributions. Unlike prior methods that regularize all paths indiscriminately, we selectively minimize the Jensen-Shannon Divergence (JSD) between the answer probability distributions $P(y|x, r_i)$ and $P(y|x, r_j)$ for every pair of selected perspectives:

\vspace{-6mm}
\begin{equation}
\mathcal{L}_{cons} = \sum_{\substack{i, j \in \mathcal{I}_{\mathrm{dyn}} \\ i < j}} (\alpha_i \cdot \alpha_j) \cdot \text{JSD}(P_i \\
|| P_j)
\end{equation}



\subsection{Student Training}
The final student objective is a weighted sum: 
\vspace{-2mm}
\begin{equation}
\mathcal{L}_{total} = \mathcal{L}_{SFT} + \lambda \mathcal{L}_{cons}
\end{equation}

By filtering out incompatible paths and reinforcing consistency among the compatible ones, this mechanism ensures the student internalizes a coherent and diverse repertoire of reasoning strategies.


\section{Experiments}
\subsection{Experimental Setup}
\textbf{Datasets.}
We evaluate MIND on two categories of benchmarks to verify task performance and generalization: (1) In-Distribution mathematical problem-solving benchmarks: MATH500 \citep{lightman2023let}, GSM8K \citep{cobbe2021trainingverifierssolvemath}, SVAMP  \citep{patel2021nlpmodelsreallyable}. (2) Out-Of-Distribution datasets commonsense reasoning benchmarks: CSQA \citep{talmor2019commonsenseqa}, StrategyQA \citep{geva2021strategyqa}, GPQA-Diamond \citep{rein2023gpqa}.

\noindent \textbf{Models and Implementation Details.} 
We employ open-source Qwen3-235B as our teacher model, selected for its state-of-the-art performance. For students, we select widely-used Qwen2.5-1.5B, Qwen2.5-7B, and Llama3.1-8B to rigorously evaluate scalability and architectural universality. Implementation settings are detailed in Table \ref{tab:hyperparameters}.



\noindent \textbf{Baselines.}
We compare MIND against a diverse set of baselines:  (1) Zero-shot CoT on base models, specifically Qwen2.5-1.5B-Instruct, Qwen2.5-7B-Instruct, and Llama3.1-8B-Instruct; (2) SbS-KD \citep{hsieh2023distilling}, representing standard vanilla CoT distillation; (3) MCC-KD \citep{chen2023mcc}, a multi-path distillation approach enforcing consistency; (4) MoDE-CoTD \citep{li2024mode}, a distillation method with mixture of decoupled experts. (5) EDIT \citep{dai-etal-2025-capture}, which emphasizes mistake-driven key reasoning steps. Additionally, we evaluate Single-Perspective Variants (students trained on individual perspectives separately) to validate the effectiveness of our dynamic fusion mechanism.

\subsection{Main Results}
Table\ref{tab:mian-result} summarizes the performance of MIND across diverse student models and benchmarks. MIND consistently outperforms the base model and surpasses all baselines by significant margins. Experiments with Llama3.1-8B and Qwen2.5-1.5B further demonstrate MIND’s robustness across different model architectures and scales. We analyze these along two key dimensions: in-distribution performance and out-of-distribution generalization.

\noindent \textbf{MIND Enhancesd In-Distribution Performance.} 
On in-distribution tasks, MIND achieves substantial gains by enabling students to adaptively internalize diverse reasoning strategies. Notably, while traditional distillation often causes parameter-constrained models (e.g., Qwen2.5-1.5B) to overfit—showing regression on tasks with slight distributional shifts—MIND maintains robust improvements comparable to the larger Qwen2.5-7B (avg. gain +3.63 / 5.10\%). This suggests that MIND facilitates the acquisition of generalized reasoning capabilities rather than the rote memorization of teacher traces.

\noindent \textbf{MIND Mitigates Catastrophic Forgetting and Boosts Generalization.} 
Unlike baseline methods that suffer from overfitting and catastrophic forgetting on knowledge-intensive OOD benchmarks (e.g., CSQA, StrategyQA), MIND achieves an average OOD accuracy improvement of 3.15 (6.41\%). Even compared to EDIT, which explicitly optimized for generalization, MIND demonstrates superior robustness in complex cross-task transfer and adaptability to small models. MIND addresses this by allowing the student to selectively assimilate only those reasoning patterns compatible with its current state, thereby minimizing interference with its pre-existing knowledge space. Remarkably, on the challenging GPQA-Diamond dataset, MIND delivers a gain of 6.8 (26.57\%). We attribute this to our perspective fusion mechanism that enables students to learn diverse cognitive styles rather than mimicking a monolithic reasoning process.

\noindent \textbf{Necessity of Multi-Perspective Fusion.} Single-perspective variants ("Ours w/o fusion") significantly underperform MIND and even lag behind baselines on OOD tasks. This degradation suggests that enforcing a rigid, monolithic reasoning mode disrupts cognitive stability, particularly in capacity-constrained models. Thus, dynamic fusion is critical for establishing a robust reasoning manifold that generalizes across domains.

\noindent \textbf{Data Efficiency.} MIND achieves these superior results using only 497 training samples, a magnitude fewer than the thousands required by standard distillation baselines. This extreme data efficiency effectively offsets the computational overhead of dynamic preference calculation, rendering the overall training process highly efficient.

\begin{table*}[t]
\centering
\scriptsize
\setlength{\tabcolsep}{4.5pt}
\renewcommand{\arraystretch}{1.1}

\definecolor{gain}{RGB}{0,128,128}

\begin{tabular}{l c c c c c c c c}
\toprule
\multicolumn{1}{c}{} &
\multicolumn{4}{c}{\textbf{In-Distribution}} &
\multicolumn{4}{c}{\textbf{Out-Of-Distribution}} \\
\cmidrule(lr){2-5}\cmidrule(lr){6-9}
\textbf{Method} &
\textbf{MATH500} & \textbf{GSM8K} & \textbf{SVAMP} & \textbf{Avg gain} &
\textbf{CSQA} & \textbf{StrategyQA} & \textbf{GPQA-D} & \textbf{Avg gain}  \\
\midrule

\rowcolor{gray!15}
\multicolumn{9}{c}{\textbf{Qwen2.5-7B}} \\
\midrule

Base (Qwen2.5-7B-Instruct) & 77.20 & 92.36 & 90.33 & \textcolor{blue}{$\uparrow$3.99} & 83.45 & 68.68 & 30.30 & \textcolor{blue}{$\uparrow$4.46} \\
SbS \citep{hsieh2023distilling} & $77.40_{\textcolor{gain}{\uparrow\,0.20}}$ & $94.77_{\textcolor{gain}{\uparrow\,2.41}}$ & $93.00_{\textcolor{gain}{\uparrow\,2.67}}$ & \textcolor{blue}{$\uparrow$2.23} & $83.20_{\textcolor{red}{\downarrow\,0.25}}$ & $67.25_{\textcolor{red}{\downarrow\,1.43}}$ & $27.46_{\textcolor{red}{\downarrow\,2.84}}$ & \textcolor{blue}{$\uparrow$5.97} \\
MCC \citep{chen2023mcc} & $82.20_{\textcolor{gain}{\uparrow\,5.00}}$ & $90.52_{\textcolor{red}{\downarrow\,1.84}}$ & $91.00_{\textcolor{gain}{\uparrow\,0.67}}$ & \textcolor{blue}{$\uparrow$2.71} & $81.72_{\textcolor{red}{\downarrow\,1.73}}$ & $67.03_{\textcolor{red}{\downarrow\,1.65}}$ & $26.77_{\textcolor{red}{\downarrow\,3.53}}$ & \textcolor{blue}{$\uparrow$6.76} \\
MoDE \citep{li2024mode} & $77.67_{\textcolor{gain}{\uparrow\,0.47}}$ & $94.16_{\textcolor{gain}{\uparrow\,1.80}}$ & $93.33_{\textcolor{gain}{\uparrow\,3.00}}$ & \textcolor{blue}{$\uparrow$2.23} & $83.70_{\textcolor{gain}{\uparrow\,0.25}}$ & $67.03_{\textcolor{red}{\downarrow\,1.65}}$ & $24.75_{\textcolor{red}{\downarrow\,5.55}}$ & \textcolor{blue}{$\uparrow$6.78} \\
EDIT \citep{dai-etal-2025-capture} & $79.50_{\textcolor{gain}{\uparrow\,2.30}}$ & $94.28_{\textcolor{gain}{\uparrow\,2.49}}$ & $93.50_{\textcolor{gain}{\uparrow\,3.17}}$ & \textcolor{blue}{$\uparrow$1.53} & $83.80_{\textcolor{gain}{\uparrow\,0.35}}$ & $67.50_{\textcolor{red}{\downarrow\,1.18}}$ & $29.10_{\textcolor{red}{\downarrow\,1.20}}$ & \textcolor{blue}{$\uparrow$5.13} \\
Ours w/o fusion Avg. & $80.60{\scriptscriptstyle \pm 0.40}$ & $92.00{\scriptscriptstyle \pm 1.33}$ & $92.33{\scriptscriptstyle \pm 0.67}$ & \textcolor{blue}{$\uparrow$2.31} & $81.05{\scriptscriptstyle \pm 0.96}$ & $66.67{\scriptscriptstyle \pm 0.42}$ & $20.20{\scriptscriptstyle \pm 1.01}$ & \textcolor{blue}{$\uparrow$9.30} \\
\rowcolor[rgb]{0.886,0.937,0.855}
Ours w/ fusion & $\textbf{82.63}_{\textcolor{gain}{\uparrow\,5.43}}$ & $\textbf{94.92}_{\textcolor{gain}{\uparrow\,2.56}}$ & $\textbf{94.31}_{\textcolor{gain}{\uparrow\,3.98}}$ & -- & $\textbf{83.98}_{\textcolor{gain}{\uparrow\,0.52}}$ & $\textbf{70.74}_{\textcolor{gain}{\uparrow\,2.06}}$ & $\textbf{41.10}_{\textcolor{gain}{\uparrow\,10.80}}$ & -- \\

\midrule

\rowcolor{gray!15}
\multicolumn{9}{c}{\textbf{Llama3.1-8B}} \\
\midrule

Base (Llama3.1-8B-Instruct) & 46.00 & 83.89 & 87.00 & \textcolor{blue}{$\uparrow$4.11} & 74.77 & 70.74 & 21.71 & \textcolor{blue}{$\uparrow$2.25} \\
SbS \citep{hsieh2023distilling} & $52.60_{\textcolor{gain}{\uparrow\,6.60}}$ & $86.96_{\textcolor{gain}{\uparrow\,3.07}}$ & $87.67_{\textcolor{gain}{\uparrow\,0.67}}$ & \textcolor{blue}{$\uparrow$0.67} & $75.02_{\textcolor{gain}{\uparrow\,0.25}}$ & $71.05_{\textcolor{gain}{\uparrow\,0.31}}$ & $19.70_{\textcolor{red}{\downarrow\,2.01}}$ & \textcolor{blue}{$\uparrow$4.23} \\
MCC \citep{chen2023mcc} & $53.00_{\textcolor{gain}{\uparrow\,7.00}}$ & $85.01_{\textcolor{gain}{\uparrow\,1.12}}$ & $83.33_{\textcolor{red}{\downarrow\,3.67}}$ & \textcolor{blue}{$\uparrow$2.63} & $73.33_{\textcolor{red}{\downarrow\,1.44}}$ & $67.99_{\textcolor{red}{\downarrow\,2.75}}$ & $18.53_{\textcolor{red}{\downarrow\,3.18}}$ & \textcolor{blue}{$\uparrow$3.26} \\
MoDE \citep{li2024mode} & $50.41_{\textcolor{gain}{\uparrow\,4.41}}$ & $84.00_{\textcolor{gain}{\uparrow\,0.11}}$ & $82.00_{\textcolor{red}{\downarrow\,5.00}}$ & \textcolor{blue}{$\uparrow$4.28} & $73.55_{\textcolor{red}{\downarrow\,1.22}}$ & $68.50_{\textcolor{red}{\downarrow\,2.24}}$ & $20.71_{\textcolor{red}{\downarrow\,1.00}}$ & \textcolor{blue}{$\uparrow$3.24} \\
EDIT \citep{dai-etal-2025-capture} & $52.80_{\textcolor{gain}{\uparrow\,6.80}}$ & $86.91_{\textcolor{gain}{\uparrow\,3.02}}$ & $87.67_{\textcolor{gain}{\uparrow\,0.67}}$ & \textcolor{blue}{$\uparrow$0.62} & $74.82_{\textcolor{gain}{\uparrow\,0.05}}$ & $71.15_{\textcolor{gain}{\uparrow\,0.41}}$ & $20.20_{\textcolor{red}{\downarrow\,1.51}}$ & \textcolor{blue}{$\uparrow$2.12} \\
Ours w/o fusion Avg. & $51.60{\scriptscriptstyle \pm 0.80}$ & $83.96{\scriptscriptstyle \pm 1.25}$ & $82.33{\scriptscriptstyle \pm 0.67}$ & \textcolor{blue}{$\uparrow$3.78} & $72.02{\scriptscriptstyle \pm 1.31}$ & $68.21{\scriptscriptstyle \pm 0.59}$ & $17.55{\scriptscriptstyle \pm 1.52}$ & \textcolor{blue}{$\uparrow$4.92} \\
\rowcolor[rgb]{0.886,0.937,0.855}
Ours w/ fusion & $\textbf{53.73}_{\textcolor{gain}{\uparrow\,7.73}}$ & $\textbf{87.51}_{\textcolor{gain}{\uparrow\,3.62}}$ & $\textbf{88.00}_{\textcolor{gain}{\uparrow\,1.00}}$ & -- & $\textbf{75.08}_{\textcolor{gain}{\uparrow\,0.31}}$ & $\textbf{72.71}_{\textcolor{gain}{\uparrow\,1.97}}$ & $\textbf{24.75}_{\textcolor{gain}{\uparrow\,3.04}}$ & -- \\

\midrule

\rowcolor{gray!15}
\multicolumn{9}{c}{\textbf{Qwen2.5-1.5B}} \\
\midrule

Base (Qwen2.5-1.5B-Instruct) & 50.40 & 72.81 & 79.33 & \textcolor{blue}{$\uparrow$3.63} & 74.20 & 58.07 & 24.75 & \textcolor{blue}{$\uparrow$2.75} \\
SbS \citep{hsieh2023distilling} & $49.80_{\textcolor{red}{\downarrow\,0.60}}$ & $73.91_{\textcolor{gain}{\uparrow\,1.10}}$ & $78.52_{\textcolor{red}{\downarrow\,0.81}}$ & \textcolor{blue}{$\uparrow$3.73} & $73.46_{\textcolor{red}{\downarrow\,0.74}}$ & $56.55_{\textcolor{red}{\downarrow\,1.52}}$ & $22.99_{\textcolor{red}{\downarrow\,1.76}}$ & \textcolor{blue}{$\uparrow$4.10} \\
MCC \citep{chen2023mcc} & $53.60_{\textcolor{gain}{\uparrow\,3.20}}$ & $70.54_{\textcolor{red}{\downarrow\,2.27}}$ & $79.80_{\textcolor{gain}{\uparrow\,0.47}}$ & \textcolor{blue}{$\uparrow$3.16} & $71.32_{\textcolor{red}{\downarrow\,2.88}}$ & $56.99_{\textcolor{red}{\downarrow\,1.08}}$ & $20.71_{\textcolor{red}{\downarrow\,4.04}}$ & \textcolor{blue}{$\uparrow$5.43} \\
MoDE \citep{li2024mode} & $52.60_{\textcolor{gain}{\uparrow\,2.20}}$ & $70.31_{\textcolor{red}{\downarrow\,2.50}}$ & $79.33_{\textcolor{gain}{\uparrow\,0.00}}$ & \textcolor{blue}{$\uparrow$3.72} & $75.14_{\textcolor{gain}{\uparrow\,0.94}}$ & $55.46_{\textcolor{red}{\downarrow\,2.61}}$ & $21.71_{\textcolor{red}{\downarrow\,3.04}}$ & \textcolor{blue}{$\uparrow$4.33} \\
EDIT \citep{dai-etal-2025-capture} & $51.20_{\textcolor{gain}{\uparrow\,0.80}}$ & $74.50_{\textcolor{gain}{\uparrow\,1.69}}$ & $79.60_{\textcolor{gain}{\uparrow\,0.27}}$ & \textcolor{blue}{$\uparrow$2.71} & $74.47_{\textcolor{gain}{\uparrow\,0.27}}$ & $57.22_{\textcolor{red}{\downarrow\,0.85}}$ & $23.74_{\textcolor{red}{\downarrow\,1.01}}$ & \textcolor{blue}{$\uparrow$3.27} \\
Ours w/o fusion Avg. & $52.00{\scriptscriptstyle \pm 0.80}$ & $70.52{\scriptscriptstyle \pm 0.36}$ & $77.66{\scriptscriptstyle \pm 1.07}$ & \textcolor{blue}{$\uparrow$4.41} & $71.57{\scriptscriptstyle \pm 0.62}$ & $53.49{\scriptscriptstyle \pm 2.11}$ & $16.67{\scriptscriptstyle \pm 3.03}$ & \textcolor{blue}{$\uparrow$7.83} \\
\rowcolor[rgb]{0.886,0.937,0.855}
Ours w/ fusion & $\textbf{54.09}_{\textcolor{gain}{\uparrow\,3.69}}$ & $\textbf{76.50}_{\textcolor{gain}{\uparrow\,3.69}}$ & $\textbf{82.83}_{\textcolor{gain}{\uparrow\,3.50}}$ & -- & $\textbf{74.77}_{\textcolor{gain}{\uparrow\,0.57}}$ & $\textbf{59.17}_{\textcolor{gain}{\uparrow\,1.10}}$ & $\textbf{31.31}_{\textcolor{gain}{\uparrow\,6.56}}$ & -- \\

\bottomrule
\end{tabular}

\caption{Performance comparison of MIND and other methods. All baselines are trained on their original settings.}
\vspace{-6mm}
\label{tab:mian-result}
\end{table*}

\subsection{Latent Space Distribution Analysis}
\label{sec:Latent Space Analysis}
To verify the internalization of diverse reasoning paradigms, we visualize the student's intrinsic reasoning manifold using the DPMM-structured encoder derived in Section \ref{sec:Latent Space Analysis}. Cognitive inclination is quantified via the Euclidean distance between projected representations and pre-identified cluster centroids in Figure \ref{fig:latent_state_visualization}(d).

\begin{figure}[t]
    \centering
    \includegraphics[width=0.8\linewidth]{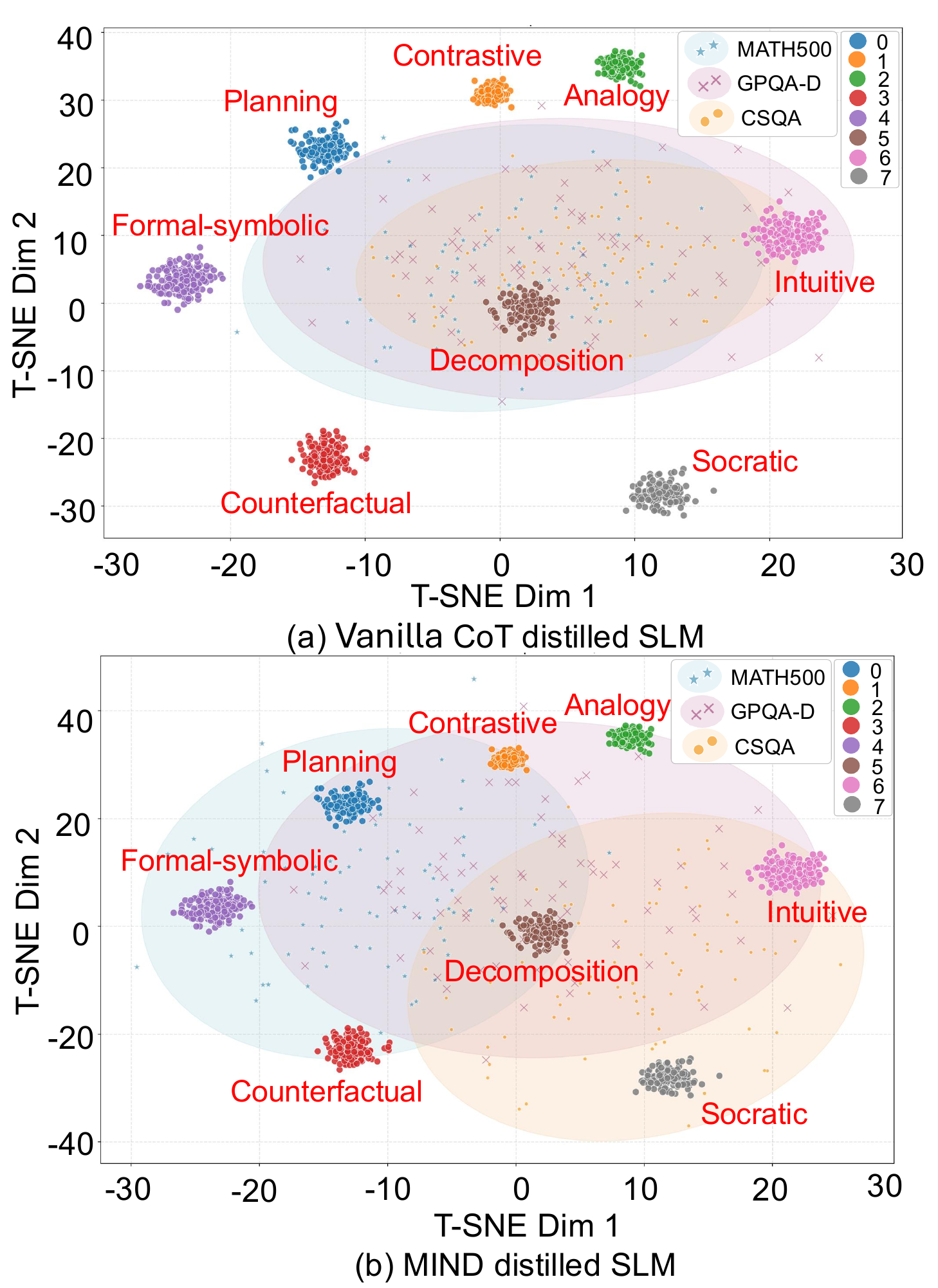}
    \vspace{-2mm}
    \caption{\textbf{Topological comparison of the student's latent reasoning manifold.} (a) The Vanilla CoT distilled student exhibits severe mode collapse. (b) The MIND-distilled student demonstrates comprehensive coverage.}
    \vspace{-6mm}
    \label{fig:reasoning_space_visualization}
\end{figure}

Figure~\ref{fig:reasoning_space_visualization} reveals a stark contrast between distillation paradigms.
The vanilla baseline (Figure~\ref{fig:reasoning_space_visualization}(a)) exhibits a decision space concentrated on narrow reasoning modes, reflecting an overfit to rigid templates and leading to brittle OOD generalization. Conversely, MIND-distilled student (Figure~\ref{fig:reasoning_space_visualization}(b)) covers the entire reasoning manifold with disentangled primitives. This broad coverage allows the model to navigate its cognitive topology, activate the most appropriate reasoning mode, or a synergistic combination tailored to specific problems.


\begin{figure}[t]
    \centering    \includegraphics[width=1\linewidth]{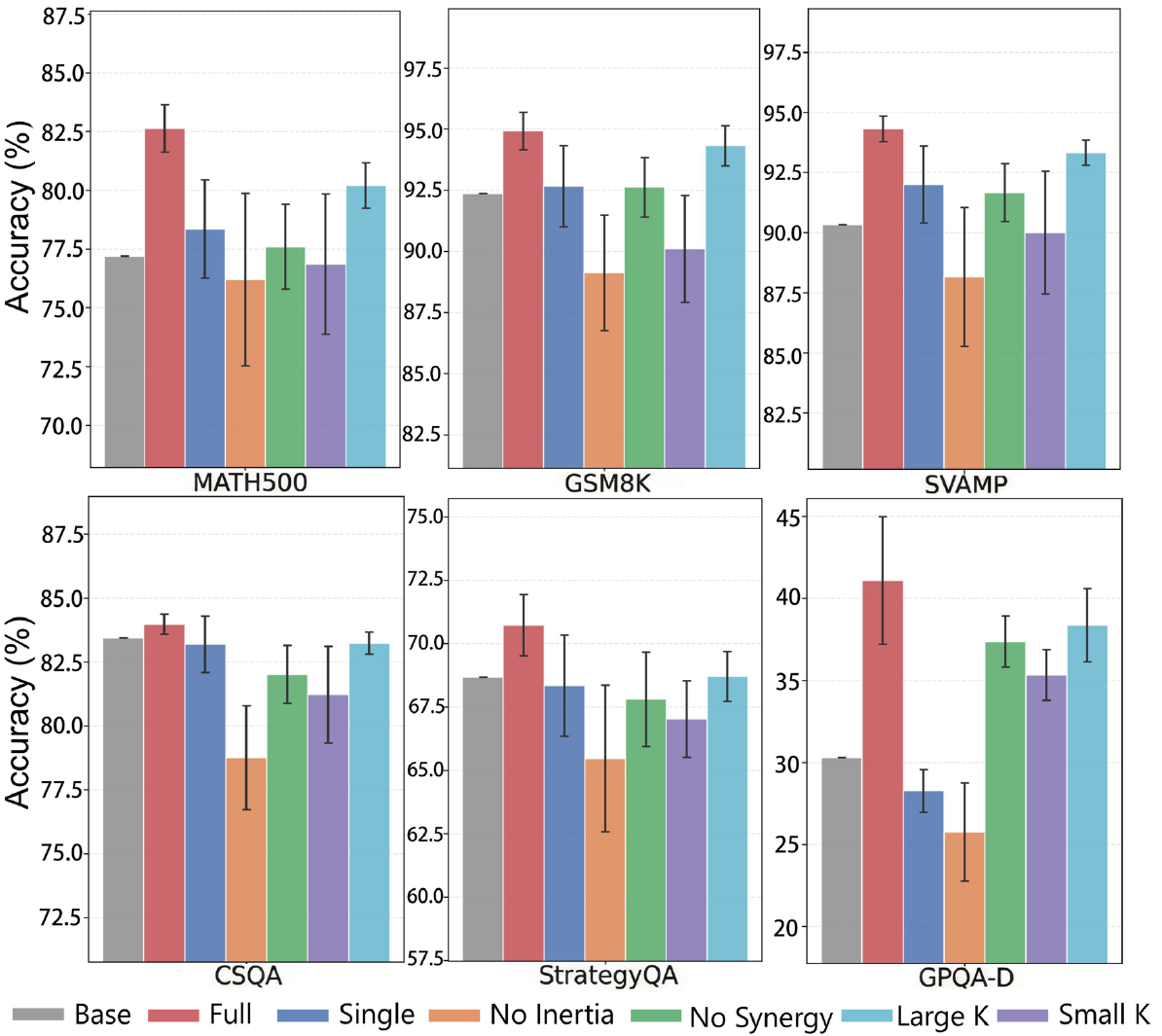}
    \caption{Comprehensive ablation study on model performance. We use top-K to replace the soft-threshold to better control the number of perspectives for clarity.}
    \vspace{-6mm}
    \label{fig:main_ablation_accuarcy}
\end{figure}

Furthermore, the specific task distributions in Figure ~\ref{fig:reasoning_space_visualization} validate the efficacy of the Synergy Layer. We observe that the model does not randomly sample perspectives but exhibits task-adaptive activation: logic-intensive tasks (e.g., MATH500) gravitate towards symbolic reasoning clusters, while semantic-intensive tasks (e.g., CSQA) shift towards intuitive ones. Notably, for the highly complex GPQA-Diamond dataset, which demands heterogeneous capabilities, the student’s representations are distributed across multiple synergistic clusters, demonstrating the model’s capacity to compose sophisticated strategies from basic primitives. These results empirically confirm that MIND equips students with a versatile cognitive landscape, directly driving superior performance in all scenarios.

\begin{figure}[t]
    \centering
    \includegraphics[width=0.8\linewidth]{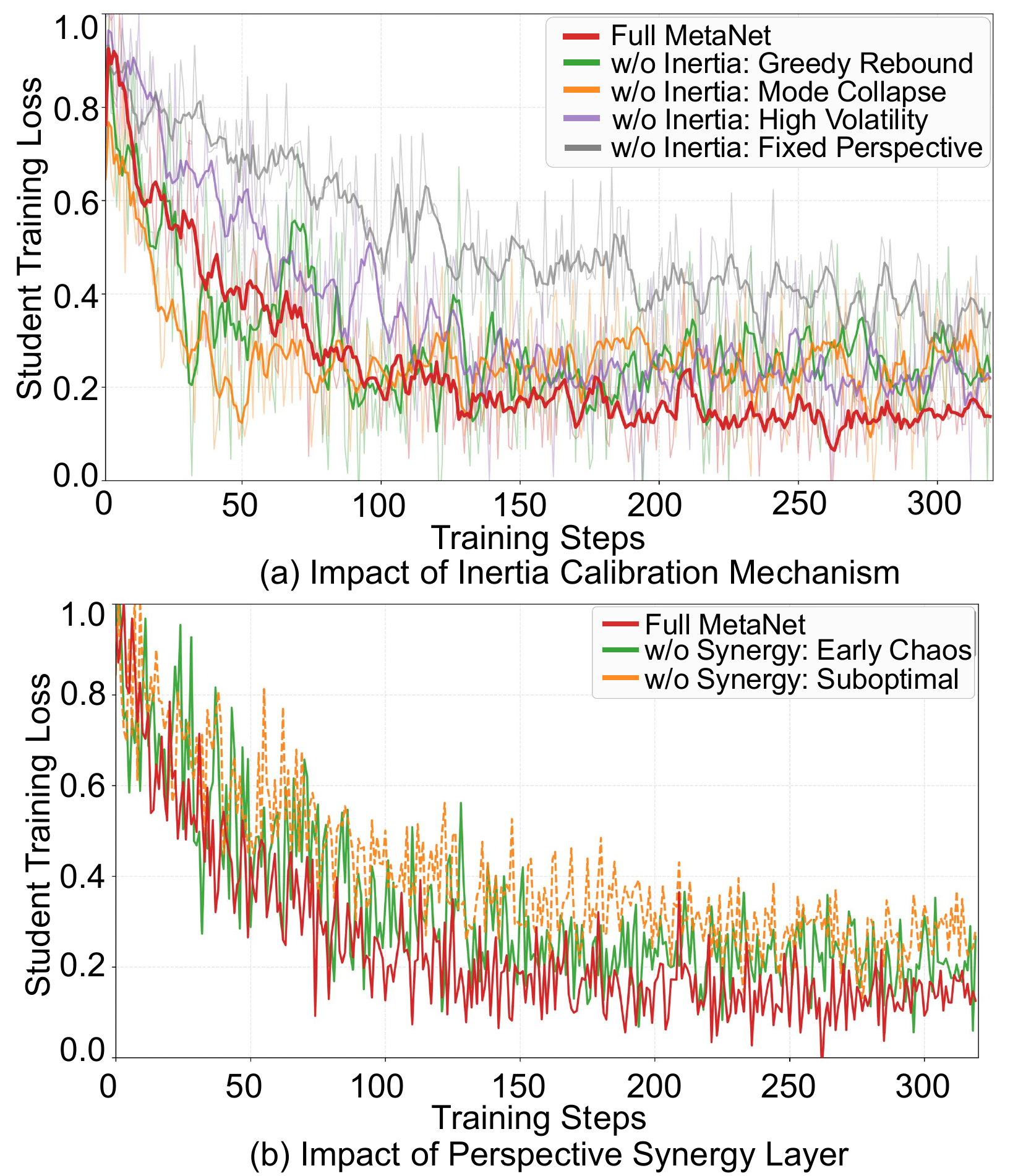}
    \vspace{-2mm}
    \caption{Effect of the Inertia Calibration Mechanism and Synergy Layer. Moving average was performed on the training loss in (a) to enhance the overall trend.}
    \vspace{-6mm}
    \label{fig:ablation_study_7_curves_2in1_virtical}
\end{figure}

\subsection{Ablation Study}

We analyze the impact of the \textit{Inertia Calibration Mechanism} and the \textit{Perspective Synergy Layer} on the student model's training stability. Comprehensive results are illustrated in Figure~\ref{fig:main_ablation_accuarcy}.

\noindent \textbf{Impact of Inertia Calibration Mechanism.} Removing the parametric inertia matrix exposes the system to instantaneous noise($\alpha_k \propto \exp(-\mathcal{L}_{real}^{(k)})$), degenerates the system into a reactive weighting scheme susceptible to high-variance aleatoric uncertainty, leading to four failure modes (Figure~\ref{fig:ablation_study_7_curves_2in1_virtical}(a)):
\textit{1) Greedy Rebound:} The model initially exploits easy samples for rapid loss reduction (greedy optimization) but suffers a sharp rebound when these shortcuts fail on complex instances, leading to unstable convergence.
\textit{2) Mode Collapse:} Premature plateauing at a sub-optimal level as the model overfits to a single "shortcut" perspective that offers low initial loss, discarding valuable reasoning paths.
\textit{3) High Volatility:} Violent weight fluctuations between mini-batches, resulting in a jagged loss trajectory that hinders optimization.
\textit{4) Fixed Perspective:} Converges to a single perspective with a smooth but slow descent on loss curve, failing to leverage curriculum-based efficiency.


\noindent \textbf{Impact of Perspective Synergy Layer.} 
The Synergy Layer aggregates cross-perspective information. Ablating this layer forces the scoring heads to evaluate paths in isolation, ignoring structural dependencies and leading to two significant degradations shown in Figure~\ref{fig:ablation_study_7_curves_2in1_virtical}(b): \textit{1) Early Chaos:} Without synergy, the model struggles to distinguish between conflicting paths, leading to high-variance noise in the early stages of training. The lack of a "soft voting" mechanism delays the formation of a stable consensus on optimal reasoning modes.
\textit{2) Suboptimal Convergence:} The inability to leverage mutual reinforcement hinders the minimization of the overall training loss, resulting in a slower convergence rate and a higher final loss plateau.

\begin{figure}[t]
    \centering
    \includegraphics[width=1\linewidth]{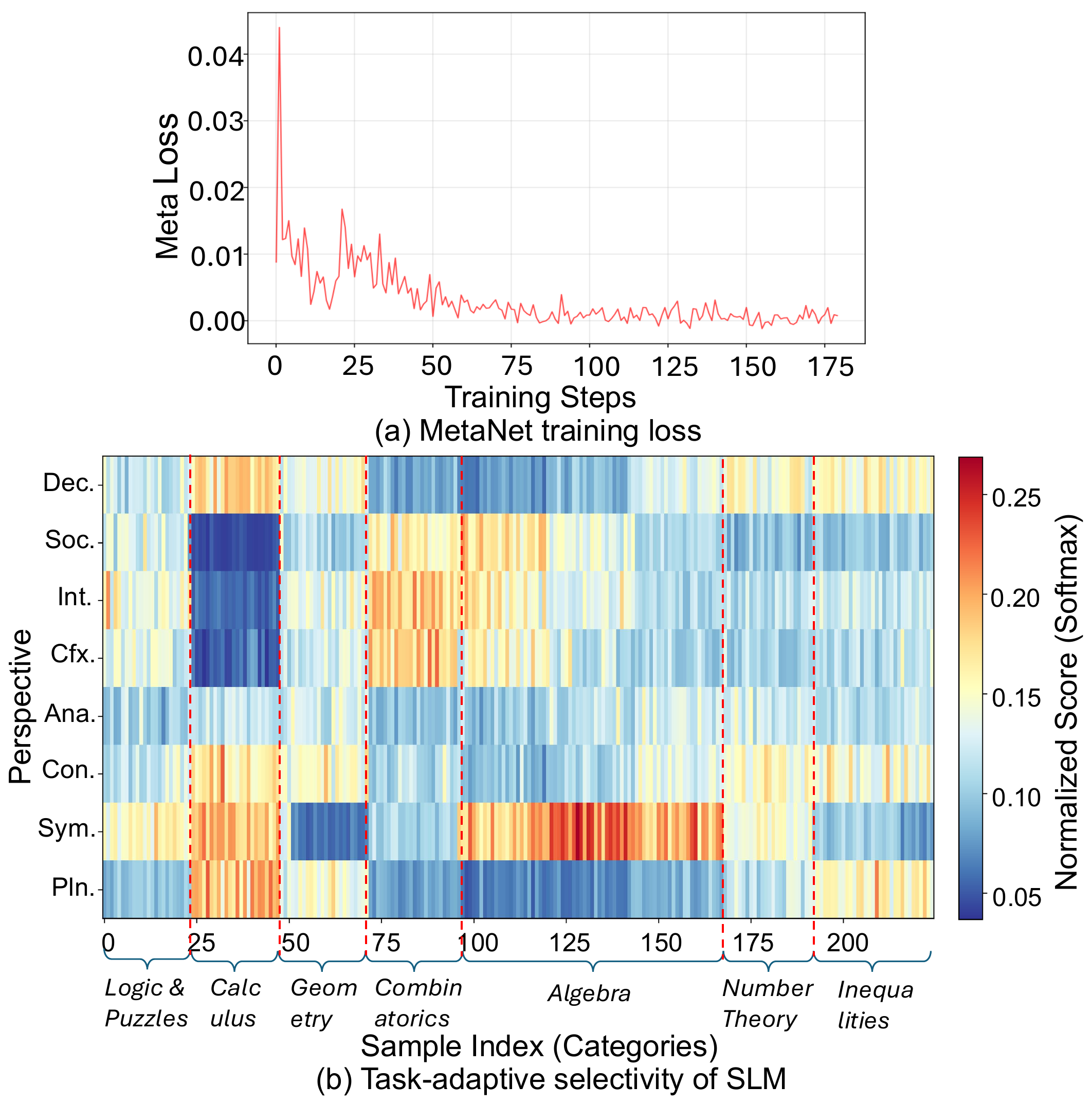}
    \vspace{-8mm}
    \caption{Evaluation to answer the question: Does the student exhibit an intrinsic, evolving preference for specific cognitive styles, or is this differentiation merely an artifact of the MetaNet's enforcement?}
    \vspace{-6mm}
    \label{fig:metaloss_scores_heatmap_2in1}
\end{figure}

\section{Analysis}
To determine whether the student's reasoning differentiation arises from intrinsic cognitive evolution or merely as an artifact of MetaNet's enforcement, we leverage MetaNet as a dynamic "cognitive probe" to monitor the training dynamics.

\noindent \textbf{Synchronized Training Dynamics.} Figure~\ref{fig:metaloss_scores_heatmap_2in1}(a) illustrates the co-evolution of the student and MetaNet. The initial volatility in MetaNet's loss reflects the student's rapid cognitive restructuring during early distillation, where MetaNet acts as a lagging indicator recalibrating to shifting representations. The subsequent stabilization confirms that the student has settled into a stable reasoning manifold, achieving high-fidelity alignment with MetaNet's weighting.

\noindent \textbf{Task-Specific Cognitive Preferences.} Figure~\ref{fig:ablation_study_7_curves_2in1_virtical}(b) validates that the student's preferences are semantically grounded. The model demonstrates task-adaptive selectivity (e.g., prioritizing spatial heuristics for geometry versus symbolic derivation for algebra). This confirms that MIND empowers the student to autonomously navigate its cognitive repertoire rather than mimicking static templates, providing a mechanistic explanation for the robust generalization observed in our experiments.

\section{Conclusion}
In this work, we proposed MIND, which transforms distillation from passive mimicry into active cognitive construction by dynamically synthesizing diverse reasoning perspectives aligned with the student's evolving capacity. MIND achieves SOTA results on ID and OOD benchmarks. Our comprehensive experiments confirm that SLMs can transcend imitation to become compact models equipped with robust, universal reasoning capabilities.

\section{Limitations}
There are two potential limitations of our work:
(1) While our eight synthesized cognitive perspectives prove highly effective for complex logical and mathematical reasoning tasks, they may not fully encompass the reasoning needs of highly subjective or creative domains (e.g., literary analysis or open-ended storytelling). Extending the MIND framework to such non-deterministic tasks would likely necessitate defining a new set of domain-specific cognitive primitives.

(2) Our current evaluation is primarily conducted on widely-used open-source student models. Future work should extend to a broader spectrum of student architectures and sizes to fully establish the universality of MIND. Additionally, validating the framework with a wider array of teacher models, including proprietary closed-source LLMs, remains an important direction to further explore the boundaries of capability transfer.

\bibliography{custom}

\begin{thebibliography}{32}
\providecommand{\natexlab}[1]{#1}

\bibitem[{Ainsworth(2006)}]{ainsworth2006deft}
Shaaron Ainsworth. 2006.
\newblock Deft: A conceptual framework for considering learning with multiple representations.
\newblock \emph{Learning and instruction}, 16(3):183--198.

\bibitem[{Brown et~al.(2020)Brown, Mann, Ryder, Subbiah, Kaplan, Dhariwal, Neelakantan, Shyam, Sastry, Askell et~al.}]{brown2020language}
Tom Brown, Benjamin Mann, Nick Ryder, Melanie Subbiah, Jared~D Kaplan, Prafulla Dhariwal, Arvind Neelakantan, Pranav Shyam, Girish Sastry, Amanda Askell, and 1 others. 2020.
\newblock Language models are few-shot learners.
\newblock In \emph{Advances in Neural Information Processing Systems (NeurIPS)}, volume~33, pages 1877--1901.

\bibitem[{Bubeck et~al.(2023)Bubeck, Chandrasekaran, Eldan, Gehrke, Horvitz, Kamar, Lee, Lee, Li, Lundberg et~al.}]{bubeck2023sparks}
S{\'e}bastien Bubeck, Varun Chandrasekaran, Ronen Eldan, Johannes Gehrke, Eric Horvitz, Ece Kamar, Peter Lee, Yin~Tat Lee, Yuanzhi Li, Scott Lundberg, and 1 others. 2023.
\newblock Sparks of artificial general intelligence: Early experiments with gpt-4.
\newblock \emph{arXiv preprint arXiv:2303.12712}.

\bibitem[{Chen et~al.(2023)Chen, Wu, Quan, Wang, Yan, and Zhang}]{chen2023mcc}
Hongzhan Chen, Siyue Wu, Xiaojun Quan, Rui Wang, Ming Yan, and Ji~Zhang. 2023.
\newblock Mcc-kd: Multi-cot consistent knowledge distillation.
\newblock \emph{arXiv preprint arXiv:2310.14747}.

\bibitem[{Chen et~al.(2025)Chen, Sun, Wenjin, Zhang, Chen, Sun, Su, Pan, Klakow, Li et~al.}]{chen2025unveiling}
Xinghao Chen, Zhijing Sun, Guo Wenjin, Miaoran Zhang, Yanjun Chen, Yirong Sun, Hui Su, Yijie Pan, Dietrich Klakow, Wenjie Li, and 1 others. 2025.
\newblock Unveiling the key factors for distilling chain-of-thought reasoning.
\newblock In \emph{Findings of the Association for Computational Linguistics: ACL 2025}, pages 15094--15119.

\bibitem[{Chowdhery et~al.(2023)Chowdhery, Narang, Devlin, Bosma, Mishra, Roberts, Barham, Chung, Sutton, Gehrmann et~al.}]{chowdhery2023palm}
Aakanksha Chowdhery, Sharan Narang, Jacob Devlin, Maarten Bosma, Gaurav Mishra, Adam Roberts, Paul Barham, Hyung~Won Chung, Charles Sutton, Sebastian Gehrmann, and 1 others. 2023.
\newblock Palm: Scaling language modeling with pathways.
\newblock \emph{Journal of Machine Learning Research}, 24(240):1--113.

\bibitem[{Cobbe et~al.(2021)Cobbe, Kosaraju, Bavarian, Chen, Jun, Kaiser, Plappert, Tworek, Hilton, Nakano, Hesse, and Schulman}]{cobbe2021trainingverifierssolvemath}
Karl Cobbe, Vineet Kosaraju, Mohammad Bavarian, Mark Chen, Heewoo Jun, Lukasz Kaiser, Matthias Plappert, Jerry Tworek, Jacob Hilton, Reiichiro Nakano, Christopher Hesse, and John Schulman. 2021.
\newblock \href {https://arxiv.org/abs/2110.14168} {Training verifiers to solve math word problems}.
\newblock \emph{Preprint}, arXiv:2110.14168.

\bibitem[{Dai et~al.(2024)Dai, Li, Zhou, and Hu}]{dai2024improve}
Chengwei Dai, Kun Li, Wei Zhou, and Songlin Hu. 2024.
\newblock Improve student's reasoning generalizability through cascading decomposed cots distillation.
\newblock \emph{arXiv preprint arXiv:2405.19842}.

\bibitem[{Dai et~al.(2025)Dai, Li, Zhou, and Hu}]{dai-etal-2025-capture}
Chengwei Dai, Kun Li, Wei Zhou, and Songlin Hu. 2025.
\newblock \href {https://doi.org/10.18653/v1/2025.acl-long.21} {Capture the key in reasoning to enhance {C}o{T} distillation generalization}.
\newblock In \emph{Proceedings of the 63rd Annual Meeting of the Association for Computational Linguistics (Volume 1: Long Papers)}, pages 441--465, Vienna, Austria. Association for Computational Linguistics.

\bibitem[{Feng et~al.(2024)Feng, Li, Chenglin, Chen, Yu, and Zhang}]{feng2024teaching}
Tao Feng, Yicheng Li, Li~Chenglin, Hao Chen, Fei Yu, and Yin Zhang. 2024.
\newblock Teaching small language models reasoning through counterfactual distillation.
\newblock In \emph{Proceedings of the 2024 Conference on Empirical Methods in Natural Language Processing}, pages 5831--5842.

\bibitem[{Fu et~al.(2023)Fu, Peng, Ou, Sabharwal, and Khot}]{fu2023specializing}
Yao Fu, Hao Peng, Litu Ou, Ashish Sabharwal, and Tushar Khot. 2023.
\newblock Specializing smaller language models towards multi-step reasoning.
\newblock In \emph{International Conference on Machine Learning}, pages 10421--10430. PMLR.

\bibitem[{Geva et~al.(2021)Geva, Khashabi, Segal, and et~al.}]{geva2021strategyqa}
Mor Geva, Daniel Khashabi, Elad Segal, and et~al. 2021.
\newblock Did aristotle use a laptop? a question answering benchmark with implicit reasoning strategies.
\newblock In \emph{Proceedings of EMNLP}.

\bibitem[{Hendrycks et~al.(2021)Hendrycks, Wang, Maziarz, Song, and ...}]{hendrycks2021measuring}
Dan Hendrycks, Collin Wang, Karina Maziarz, Dawn Song, and ... 2021.
\newblock Measuring mathematical problem solving with the math dataset.
\newblock In \emph{Advances in Neural Information Processing Systems}.

\bibitem[{Ho et~al.(2022)Ho, Schmid, and Yun}]{Ho2022LargeLM}
Namgyu Ho, Laura Schmid, and Se-Young Yun. 2022.
\newblock Large language models are reasoning teachers.
\newblock In \emph{Proceedings of the 60th Annual Meeting of the Association for Computational Linguistics (ACL)}.

\bibitem[{Hsieh et~al.(2023)Hsieh, Li, Yeh, Nakhost, Fujii, Ratner, Krishna, Lee, and Pfister}]{hsieh2023distilling}
Cheng-Yu Hsieh, Chun-Liang Li, Chih-Kuan Yeh, Hootan Nakhost, Yasuhisa Fujii, Alex Ratner, Ranjay Krishna, Chen-Yu Lee, and Tomas Pfister. 2023.
\newblock Distilling step-by-step! outperforming larger language models with less training data and smaller model sizes.
\newblock In \emph{Findings of the Association for Computational Linguistics: ACL 2023}, pages 8003--8017.

\bibitem[{Imani et~al.(2023)Imani, Du, and Shrivastava}]{imani2023math}
Shima Imani, Liang Du, and Harsh Shrivastava. 2023.
\newblock Math-prompter: Mathematical reasoning using large language models.
\newblock \emph{arXiv preprint arXiv:2303.05398}.

\bibitem[{Jiang et~al.(2025)Jiang, Lu, Lin, Han, and Sun}]{jiang2025teach}
Wangyi Jiang, Yaojie Lu, Hongyu Lin, Xianpei Han, and Le~Sun. 2025.
\newblock Teach small models to reason by curriculum distillation.
\newblock In \emph{Proceedings of the 2025 Conference on Empirical Methods in Natural Language Processing}, pages 7423--7433.

\bibitem[{Kojima et~al.(2022)Kojima, Gu, Reid, Matsuo, and Iwasawa}]{kojima2022large}
Takeshi Kojima, Shixiang~Shane Gu, Machel Reid, Yutaka Matsuo, and Yusuke Iwasawa. 2022.
\newblock Large language models are zero-shot reasoners.
\newblock In \emph{Advances in Neural Information Processing Systems (NeurIPS)}, volume~35, pages 22199--22213.

\bibitem[{Li et~al.(2024{\natexlab{a}})Li, Chen, Li, Wang, Tao, Li, Chen, and Zhang}]{chenglin2024mixed}
Chenglin Li, Qianglong Chen, Liangyue Li, Caiyu Wang, Feng Tao, Yicheng Li, Zulong Chen, and Yin Zhang. 2024{\natexlab{a}}.
\newblock Mixed distillation helps smaller language models reason better.
\newblock In \emph{Findings of the Association for Computational Linguistics: EMNLP 2024}, pages 1673--1690.

\bibitem[{Li et~al.(2024{\natexlab{b}})Li, He, Wu, Yang, Xu, jun Jun, Liu, Liu, and Zhao}]{li2024mode}
Xiang Li, Shizhu He, Jiayu Wu, Zhao Yang, Yao Xu, Yang jun Jun, Haifeng Liu, Kang Liu, and Jun Zhao. 2024{\natexlab{b}}.
\newblock Mode-cotd: Chain-of-thought distillation for complex reasoning tasks with mixture of decoupled lora-experts.
\newblock In \emph{Proceedings of the 2024 Joint International Conference on Computational Linguistics, Language Resources and Evaluation (LREC-COLING 2024)}, pages 11475--11485.

\bibitem[{Lightman et~al.(2023)Lightman, Kosaraju, Burda, Edwards, Baker, Lee, Leike, Schulman, Sutskever, and Cobbe}]{lightman2023let}
Hunter Lightman, Vineet Kosaraju, Yuri Burda, Harrison Edwards, Bowen Baker, Teddy Lee, Jan Leike, John Schulman, Ilya Sutskever, and Karl Cobbe. 2023.
\newblock Let's verify step by step.
\newblock In \emph{The Twelfth International Conference on Learning Representations}.

\bibitem[{Lin et~al.(2025)Lin, Yao, Hsu, Xie, Shuai, and Cheng}]{lin2025perspective}
Jhe-Hao Lin, Yi~Yao, Chan-Feng Hsu, Hong-Xia Xie, Hong-Han Shuai, and Wen-Huang Cheng. 2025.
\newblock Perspective-aware teaching: Adapting knowledge for heterogeneous distillation.
\newblock In \emph{Proceedings of the IEEE/CVF International Conference on Computer Vision}, pages 4178--4187.

\bibitem[{Magister et~al.(2023)Magister, Mallinson, Adamek, Malmi, and Severyn}]{magister2023teaching}
Lucie~Charlotte Magister, Jonathan Mallinson, Jakub Adamek, Eric Malmi, and Aliaksei Severyn. 2023.
\newblock Teaching small language models to reason.
\newblock In \emph{Proceedings of the 61st annual meeting of the association for computational linguistics (volume 2: short papers)}, pages 1773--1781.

\bibitem[{Nye et~al.(2021)Nye, Andreassen, Gur-Ari, Michalewski, Austin, Bieber, Dohan, Lewkowycz, Bosma, Luan et~al.}]{nye2021show}
Maxwell Nye, Anders~Johan Andreassen, Guy Gur-Ari, Henryk Michalewski, Jacob Austin, David Bieber, David Dohan, Aitor Lewkowycz, Maarten Bosma, David Luan, and 1 others. 2021.
\newblock Show your work: Scratchpads for intermediate computation with language models.
\newblock In \emph{Deep Learning for Code Workshop at NeurIPS}.

\bibitem[{Patel et~al.(2021)Patel, Bhattamishra, and Goyal}]{patel2021nlpmodelsreallyable}
Arkil Patel, Satwik Bhattamishra, and Navin Goyal. 2021.
\newblock \href {https://arxiv.org/abs/2103.07191} {Are nlp models really able to solve simple math word problems?}
\newblock \emph{Preprint}, arXiv:2103.07191.

\bibitem[{Rein et~al.(2023)Rein, Balsiger, and et~al.}]{rein2023gpqa}
Peter Rein, Fabian Balsiger, and et~al. 2023.
\newblock Gpqa: A graduate-level google-proof q\&a benchmark.
\newblock \emph{arXiv preprint arXiv:2311.12022}.

\bibitem[{Talmor et~al.(2019)Talmor, Herzig, Lourie, and Berant}]{talmor2019commonsenseqa}
Alon Talmor, Jonathan Herzig, Nicholas Lourie, and Jonathan Berant. 2019.
\newblock Commonsenseqa: A question answering challenge targeting commonsense knowledge.
\newblock In \emph{Proceedings of NAACL-HLT}.

\bibitem[{Wang et~al.(2023{\natexlab{a}})Wang, Wang, Li, Gao, Yin, and Ren}]{wang2023scott}
Peifeng Wang, Zhengyang Wang, Zheng Li, Yifan Gao, Bing Yin, and Xiang Ren. 2023{\natexlab{a}}.
\newblock Scott: Self-consistent chain-of-thought distillation.
\newblock \emph{arXiv preprint arXiv:2305.01879}.

\bibitem[{Wang et~al.(2025)Wang, Li, Xu, Cai, Song, Qi, Zhou, Huang, Wang, and Xiao}]{wang2025qcrd}
Wei Wang, Zhaowei Li, Qi~Xu, Yiqing Cai, Hang Song, Qi~Qi, Ran Zhou, Zhida Huang, Tao Wang, and Li~Xiao. 2025.
\newblock {QCRD}: Quality-guided contrastive rationale distillation for large language models.
\newblock In \emph{Proceedings of the 2025 Conference on Empirical Methods in Natural Language Processing}, pages 14345--14356.

\bibitem[{Wang et~al.(2023{\natexlab{b}})Wang, Wei, Schuurmans, Le, Chi, Narang, Chowdhery, and Zhou}]{wang2023selfconsistency}
Xuezhi Wang, Jason Wei, Dale Schuurmans, Quoc Le, Ed~Chi, Sharan Narang, Aakanksha Chowdhery, and Denny Zhou. 2023{\natexlab{b}}.
\newblock Self-consistency improves chain of thought reasoning in language models.
\newblock In \emph{International Conference on Learning Representations (ICLR)}.

\bibitem[{Wei et~al.(2022{\natexlab{a}})Wei, Tay, Bommasani, Raffel, Zoph, Borgeaud, Yogatama, Bosma, Zhou, Metzler et~al.}]{wei2022emergent}
Jason Wei, Yi~Tay, Rishi Bommasani, Colin Raffel, Barret Zoph, Sebastian Borgeaud, Dani Yogatama, Maarten Bosma, Denny Zhou, Donald Metzler, and 1 others. 2022{\natexlab{a}}.
\newblock Emergent abilities of large language models.
\newblock \emph{arXiv preprint arXiv:2206.07682}.

\bibitem[{Wei et~al.(2022{\natexlab{b}})Wei, Wang, Schuurmans, Bosma, Xia, Chi, Le, and Zhou}]{wei2022chain}
Jason Wei, Xuezhi Wang, Dale Schuurmans, Maarten Bosma, Fei Xia, Ed~Chi, Quoc~V Le, and Denny Zhou. 2022{\natexlab{b}}.
\newblock Chain-of-thought prompting elicits reasoning in large language models.
\newblock In \emph{Advances in Neural Information Processing Systems (NeurIPS)}, volume~35, pages 24824--24837.

\end{thebibliography}

\appendix
\raggedbottom

\setlength{\textfloatsep}{8pt}
\setlength{\floatsep}{6pt}
\setlength{\intextsep}{8pt}

\section{Appendix}
\label{sec:appendix}

\subsection{Details of Tasks and Datasets}

We select MATH500, GSM8K, SVAMP, CommonsenseQA, StrategyQA, and GPQA-Diamond to systematically evaluate model performance across two dimensions: mathematical reasoning and commonsense/knowledge reasoning.
The basic statistics of these benchmarks are presented in
Tables~\ref{tab:math500_subject_stats}--\ref{tab:gpqa_diamond_domain_stats}.

\vspace{0.5em}

\textbf{MATH500.}MATH500 comprises 500 problems randomly sampled from the MATH dataset~\citep{hendrycks2021measuring}, covering seven subjects across five difficulty levels.

\textbf{GSM8K.} GSM8K \citep{cobbe2021trainingverifierssolvemath} consists of approximately 8.5K high-quality, linguistically diverse grade school math word problems.

\textbf{SVAMP.} SVAMP \citep{patel2021nlpmodelsreallyable} includes 1,000 one-unknown arithmetic word problems (up to grade 4), constructed by applying structural variations to problem statements.

\textbf{CommonsenseQA.} CommonsenseQA~\citep{talmor2019commonsenseqa} contains 12,247 examples testing commonsense knowledge.

\textbf{StrategyQA.} StrategyQA~\citep{geva2021strategyqa} comprises 2,780 questions requiring multi-step strategy inference to answer questions with implicit reasoning steps.

\textbf{GPQA-Diamond.} GPQA-Diamond~\citep{rein2023gpqa} contains 198 expert-written questions in biology, physics, and chemistry, selected for high discrimination between experts and non-experts.

\vspace{0.5em} 


\begin{table}[!htbp]
\centering
\small
\renewcommand{\arraystretch}{1.15}
\setlength{\tabcolsep}{10pt}
\begin{tabular}{lcc}
\hline
\textbf{Subject Area} & \textbf{Size} & \textbf{Proportion} \\
\hline
Algebra                  & 124 & 24.8\% \\
Counting \& Probability  &  38 &  7.6\% \\
Geometry                 &  41 &  8.2\% \\
Intermediate Algebra     &  97 & 19.4\% \\
Number Theory            &  62 & 12.4\% \\
Prealgebra               &  82 & 16.4\% \\
Precalculus              &  56 & 11.2\% \\
\hline
\textbf{Total}           & \textbf{500} & \textbf{100.0\%} \\
\hline
\end{tabular}
\caption{Subject-area distribution of MATH500.}
\label{tab:math500_subject_stats}
\end{table}

\begin{table}[!htbp]
\centering
\small
\renewcommand{\arraystretch}{1.15}
\setlength{\tabcolsep}{10pt}
\begin{tabular}{lcc}
\hline
\textbf{Difficulty Level} & \textbf{Size} & \textbf{Proportion} \\
\hline
Level 1 &  43 &  8.6\% \\
Level 2 &  90 & 18.0\% \\
Level 3 & 105 & 21.0\% \\
Level 4 & 128 & 25.6\% \\
Level 5 & 134 & 26.8\% \\
\hline
\textbf{Total} & \textbf{500} & \textbf{100.0\%} \\
\hline
\end{tabular}
\caption{Difficulty-level distribution of MATH500.}
\label{tab:math500_difficulty_stats}
\end{table}

\begin{table}[!htbp]
\centering
\small
\renewcommand{\arraystretch}{1.15}
\setlength{\tabcolsep}{10pt}
\begin{tabular}{lcc}
\hline
\textbf{Operation-type} & \textbf{Size} & \textbf{Proportion} \\
\hline
Subtraction       & 160 & 53.33\% \\
Addition          &  59 & 19.67\% \\
Common-Division   &  48 & 16.00\% \\
Multiplication    &  33 & 11.00\% \\
\hline
\textbf{Total}    & \textbf{300} & \textbf{100.00\%} \\
\hline
\end{tabular}
\caption{Operation-type distribution of the SVAMP test set.}
\label{tab:svamp_test_type_stats}
\end{table}

\begin{table}[!htbp]
\centering
\small
\renewcommand{\arraystretch}{1.15}
\setlength{\tabcolsep}{10pt}
\begin{tabular}{lcc}
\hline
\textbf{Domain} & \textbf{Size} & \textbf{Proportion} \\
\hline
Chemistry &  93 & 46.97\% \\
Physics   &  86 & 43.43\% \\
Biology   &  19 &  9.60\% \\
\hline
\textbf{Total} & \textbf{198} & \textbf{100.00\%} \\
\hline
\end{tabular}
\caption{Domain distribution of the GPQA-Diamond dataset.  }
\label{tab:gpqa_diamond_domain_stats}
\end{table}


\subsection{Multi-perspective Dataset}


We provide a detailed statistical overview of the multi-perspective dataset here. Using Qwen3-235B and eight perspective prompts (Appendix~\ref{app:prompts}), we curated 497 samples from the MATH dataset, strictly following the main text's filtering and stratification criteria. Each sample contains eight valid CoT rationales yielding correct answers. Tables~\ref{tab:math_subject_stats} and~\ref{tab:math_difficulty_stats} summarize the dataset distribution by subject area and difficulty.

\vspace{0.5em} 

\begin{table}[!htbp]
\centering
\small
\renewcommand{\arraystretch}{1.15}
\setlength{\tabcolsep}{10pt}
\begin{tabular}{lcc}
\hline
\textbf{Subject Area} & \textbf{Size} & \textbf{Proportion} \\
\hline
algebra                    & 187 & 37.6\% \\
counting\_and\_probability  &  31 &  6.2\% \\
geometry                   &  25 &  5.0\% \\
intermediate\_algebra       &  43 &  8.7\% \\
number\_theory              &  59 & 11.9\% \\
prealgebra                 & 122 & 24.5\% \\
precalculus                &  30 &  6.0\% \\
\hline
\textbf{Total}             & \textbf{497} & \textbf{100.0\%} \\
\hline
\end{tabular}
\caption{Subject-area distribution of the constructed MATH-derived dataset.}
\label{tab:math_subject_stats}
\end{table}

\begin{table}[!htbp]
\centering
\small
\renewcommand{\arraystretch}{1.15}
\setlength{\tabcolsep}{10pt}
\begin{tabular}{lcc}
\hline
\textbf{Difficulty Level} & \textbf{Size} & \textbf{Proportion} \\
\hline
Level 1 &  74 & 14.9\% \\
Level 2 & 137 & 27.6\% \\
Level 3 & 124 & 24.9\% \\
Level 4 & 112 & 22.5\% \\
Level 5 &  50 & 10.1\% \\
\hline
\textbf{Total} & \textbf{497} & \textbf{100.0\%} \\
\hline
\end{tabular}
\caption{Difficulty-level distribution of the constructed MATH-derived dataset.}
\label{tab:math_difficulty_stats}
\end{table}


\subsection{Experimental Environment}
\begin{table}[ht]
\centering
\small
\begin{tabular}{ll}
\toprule
\textbf{Parameter} & \textbf{Value} \\
\midrule
\textbf{General Settings} & \\
Optimizer & AdamW \\
LoRA Target Modules & All linear layers (Attn. \& FFN) \\
Batch Size (per Device) & 4 \\
Gradient Accumulation & 4 \\
Hardware & 2 $\times$ NVIDIA A100 (80GB) \\
\midrule
\textbf{SLMs Optimization} & \\
Learning Rate (1.5B) & $5 \times 10^{-5}$ \\
Learning Rate (7B/8B) & $1 \times 10^{-4}$ \\
Training Epochs (1.5B) & 8 \\
Training Epochs (7B/8B) &  6 \\
\midrule
\textbf{MetaNet Configuration} & \\
Warmup Stage & 1 Epoch @ $1 \times 10^{-4}$ \\
Calibration Stage LR & $5 \times 10^{-5}$ \\
\bottomrule
\end{tabular}
\caption{Hyperparameter settings and hardware configuration for MIND distillation.}
\label{tab:hyperparameters}
\end{table}

For brevity, the primary hyperparameter settings and fine-tuning configurations employed in our experiments are listed in Table \ref{tab:hyperparameters}.

\subsection{Multi-perspective Prompts}
\label{app:prompts}

To encapsulate the broad reasoning manifold and mitigate the bias of any single reasoning mode, we elicited diverse CoT rationales by simulating eight teacher "stylistic signatures." By employing the prompts listed in Table~\ref{tab:reasoning_perspectives}, we tasked Qwen3-235B with constructing a multi-perspective dataset. This diversity ensures that the student model is exposed to a rich spectrum of cognitive dimensions, fostering superior generalization across complex tasks.

\begin{table*}[!t]
\centering
\small
\begin{tabular}{p{2.2cm} p{10.5cm}}
\toprule
\textbf{Perspective} & \textbf{Description} \\
\midrule
Formal Symbolic (Sym.) &
Solve the problem using strict formal reasoning. Translate all relevant information into symbolic or logical representations and follow a sequence of explicit, verifiable deductions. Intuitive leaps are avoided in favor of equations, logic operators, or formal inference rules. The final answer is presented after the last formal step. \\
\midrule
Intuitive (Int.) &
Solve the problem using intuitive and heuristic reasoning. Concepts are explained in simple, human-like terms, relying on everyday experience and plausible expectations. Heavy formalism is avoided in favor of natural understanding. The final answer is stated after the intuitive reasoning process. \\
\midrule
Decomposition (Dec.) &
Break the problem into small, concrete sub-goals. Each step explicitly addresses a component of the problem and incrementally leads toward the solution. Reasoning steps are numbered (e.g., Step~1, Step~2, \dots) without skipping intermediate logic. The final answer is provided at the end. \\
\midrule
Planning (Pln.) &
Begin with a high-level plan outlining the strategy for solving the problem. Each component of the plan is then expanded into detailed reasoning. The structure is made explicit in the order of \textit{Plan}, \textit{Execution}, and \textit{Answer}. The final answer follows the completed breakdown. \\
\midrule
Analogy (Ana.) &
Solve the problem by constructing an analogy to a simpler or more familiar situation. The analogy is explained clearly, and each of its elements is mapped back to the original problem. The inferred solution from the analogy leads to the final answer. \\
\midrule
Socratic (Soc.) &
Employ a self-questioning approach to guide reasoning. At each step, guiding questions (e.g., ``What is known?'', ``What follows?'', ``Why?'') are posed and explicitly answered. This iterative questioning drives the reasoning process, culminating in the final answer. \\
\midrule
Contrastive (Con.) &
Solve the problem by comparing multiple candidate explanations or answers. Each option is analyzed in terms of its plausibility, strengths, and weaknesses. Through contrast and elimination, the most appropriate answer is identified and stated. \\
\midrule
Counterfactual (Cfx.) &
Apply counterfactual reasoning by considering how the outcome would change if certain conditions were altered. Alternative scenarios are analyzed to reveal critical dependencies. These insights are then used to determine the correct answer under the actual conditions. The final answer is stated accordingly. \\
\bottomrule
\end{tabular}
\caption{Prompt templates of eight reasoning perspectives used in the multi-perspective dataset construction.}
\label{tab:reasoning_perspectives}
\end{table*}

\end{document}